\documentclass[conference]{IEEEtran}
\IEEEoverridecommandlockouts
\usepackage{cite}
\usepackage{amsmath,amssymb,amsfonts}
\usepackage{algorithmic}
\usepackage{graphicx}
\usepackage{textcomp}
\usepackage{xcolor}
\usepackage{makecell}
\usepackage{multirow}
\usepackage{array}
\usepackage{tabularx}
\usepackage{enumitem}
\usepackage[most]{tcolorbox}
\usepackage[table]{xcolor}
\usepackage{booktabs}
\usepackage{url}

\def\BibTeX{{\rm B\kern-.05em{\sc i\kern-.025em b}\kern-.08em
    T\kern-.1667em\lower.7ex\hbox{E}\kern-.125emX}}
\begin{document}

\title{MatPhaseBench: A Semantics-Guided Benchmark for Materials Phase Diagrams Understanding\\

\thanks{This work was supported by the National Key Research and Development Program of China under Grant No. 2025YFE0102600.}
}

% \author{\IEEEauthorblockN{Hanwen Wang}
% \IEEEauthorblockA{
% \textit{Computer Network Information Center, Chinese Academy of Sciences}\\
% \textit{University of Chinese Academy of Sciences} \\
% Beijing, China\\
% hwwang@cnic.cn}
% \and
% \IEEEauthorblockN{Sihan Liang}
% \IEEEauthorblockA{
% \textit{Computer Network Information Center, Chinese Academy of Sciences} \\
% \textit{University of Chinese Academy of Sciences} \\
% Beijing, China\\
% shliang@cnic.cn}
% \and
% \IEEEauthorblockN{Zhiwei Liu*}
% \IEEEauthorblockA{
% \textit{University of Manchester} \\
% Manchester, UK \\
% *Corresponding author: zhiweiliu0810@gmail.com}
% \and
% \IEEEauthorblockN{Yangang Wang*}
% \IEEEauthorblockA{
% \textit{Computer Network Information Center, Chinese Academy of Sciences} \\
% \textit{University of Chinese Academy of Sciences} \\
% Beijing, China\\
% *Corresponding author: wangyg@sccas.cn}
% \and
% \IEEEauthorblockN{Wei Yan}
% \IEEEauthorblockA{
% \textit{Institute of Metal Research, Chinese Academy of Sciences} \\
% Shenyang, China\\
% weiyan@imr.ac.cn}
% \and
% \IEEEauthorblockN{Yuqin Liu}
% \IEEEauthorblockA{
% \textit{China University of Geosciences} \\
% Beijing, China\\
% liuyuqin@cugb.edu.cn}
% \and
% \IEEEauthorblockN{Zongguo Wang*}
% \IEEEauthorblockA{
% \textit{Computer Network Information Center, Chinese Academy of Sciences} \\
% \textit{University of Chinese Academy of Sciences} \\
% Beijing, China\\
% *Corresponding author: wangzg@cnic.cn, 0000-0002-7719-761X}
% }

\author{
\footnotesize
\renewcommand{\arraystretch}{1.2} % 调整整行高度
\setlength{\tabcolsep}{15pt}      % 调整列间距
\begin{tabular}{c c c}

% ---------- Row 1 ----------
\begin{minipage}[t]{0.30\textwidth}
\centering
\textbf{Hanwen Wang}\\
\textit{Computer Network Information Center,}\\
\textit{Chinese Academy of Sciences}\\
\textit{University of Chinese Academy of Sciences}\\
Beijing, China\\
hwwang@cnic.cn
\end{minipage}
&
\begin{minipage}[t]{0.30\textwidth}
\centering
\textbf{Sihan Liang}\\
\textit{Computer Network Information Center,}\\
\textit{Chinese Academy of Sciences}\\
\textit{University of Chinese Academy of Sciences}\\
Beijing, China\\
shliang@cnic.cn
\end{minipage}
&
\begin{minipage}[t]{0.30\textwidth}
\centering
\textbf{Zhiwei Liu*}\\
\textit{University of Manchester}\\
Manchester, UK\\
*Corresponding author:\\
zhiweiliu0810@gmail.com
\end{minipage}
\\[7em] % 第一行和第二行间距

% ---------- Row 2 ----------
\begin{minipage}[t]{0.30\textwidth}
\centering
\textbf{Yangang Wang*}\\
\textit{Computer Network Information Center,}\\
\textit{Chinese Academy of Sciences}\\
\textit{University of Chinese Academy of Sciences}\\
Beijing, China\\
*Corresponding author:\\
wangyg@sccas.cn
\end{minipage}
&
\begin{minipage}[t]{0.30\textwidth}
\centering
\textbf{Wei Yan}\\
\textit{Institute of Metal Research,}\\
\textit{Chinese Academy of Sciences}\\
Shenyang, China\\
weiyan@imr.ac.cn
\end{minipage}
&
\begin{minipage}[t]{0.30\textwidth}
\centering
\textbf{Yuqin Liu}\\
\textit{China University of Geosciences}\\
Beijing, China\\
liuyuqin@cugb.edu.cn
\end{minipage}
\\[8em] % 第二行和第三行间距

% ---------- Row 3 ----------
\begin{minipage}[t]{0.30\textwidth}
\centering
\textbf{Zongguo Wang*}\\
\textit{Computer Network Information Center,}\\
\textit{Chinese Academy of Sciences}\\
\textit{University of Chinese Academy of Sciences}\\
Beijing, China\\
*Corresponding author:\\
wangzg@cnic.cn\\
ORCID: 0000-0002-7719-761X
\end{minipage}
&
&
\end{tabular}

}

\maketitle

\begin{abstract}
Materials phase diagrams are a core knowledge representation in materials science, encoding temperature, composition, phase stability, and phase transformation pathways, with their full understanding requiring thermodynamic mechanism analysis and scientific reasoning. Although VLMs have shown promise in scientific image understanding, their systematic evaluation on such logically complex images demanding deep mechanistic interpretation remains limited, and phase diagrams provide a challenging testbed for this purpose. We introduce MatPhaseBench, a high-quality, high-reliability benchmark for complex scientific image understanding, focused on materials phase diagrams. MatPhaseBench is constructed from 3,681 papers in classical materials science journals, from which 200 high-quality diagram-text pairs were selected, covering 189 material systems and 70 elements. The benchmark has three key features: (1) targeting complex scientific image understanding—it moves beyond simple objective tests to open-ended tasks requiring deep comprehension; (2) comprehensive image-text alignment—semantic information associated with images is fully preserved during literature mining and matching; (3) high-quality human-supervised text acquisition—all descriptions undergo strict manual validation. Experimental results show that current VLMs remain substantially behind expert-level understanding: they are largely limited to surface visual perception, lack deep reasoning grounded in thermodynamic mechanisms, have limited domain awareness and expert analytical experience, and perform poorly in distinguishing fine-grained differences in composite or multi-diagram settings. Overall, MatPhaseBench constitutes a challenging research-grade benchmark, providing a foundational platform for complex scientific image understanding, phase diagram analysis, and trustworthy multimodal AI in science. The code and datasets are available at
\url{https://github.com/Davidwhw/MatPhaseBench}.
\end{abstract}


\begin{figure}[htbp]
\centering
\includegraphics[
    width=0.95\columnwidth, 
    trim=3cm 3cm 3cm 2.5cm,
    clip
]{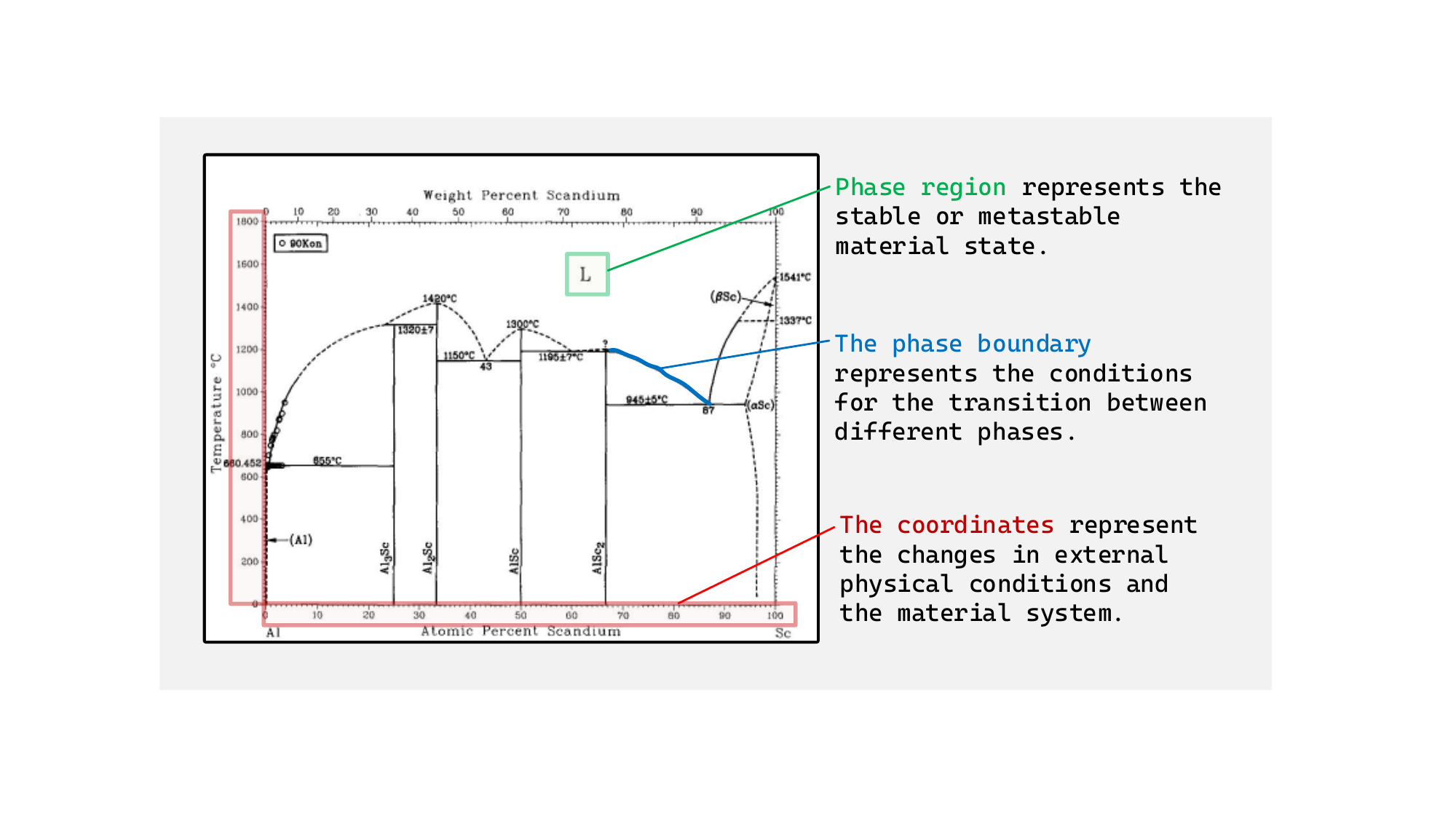}
\caption{An example of material phase diagram.}
\label{fig:phase_example}
\end{figure}

\section{Introduction}
Materials phase diagrams, as shown in Fig.~\ref{fig:phase_example}\cite{b1}, are core scientific images representing materials systems under variables such as temperature, pressure, and composition. They convey stable phase regions, phase boundaries, transformations, and invariant reactions. Experts use phase diagrams to infer phase stability, transformation pathways, and thermodynamic consistency. Their interpretation requires not only visual recognition but also understanding axes, composition ranges, phase boundaries, reaction points, and materials-science concepts, making them a suitable testbed for evaluating whether vision-language models (VLMs) can achieve complex scientific image understanding beyond surface-level perception.

Benchmarks, from ImageNet\cite{b2} and GLUE\cite{b3} to VQA\cite{b4} and MMMU\cite{b5}, are crucial for shaping AI capabilities, and their construction has increasingly shifted from general-domain image understanding to scientific and professional scenarios with the rise of VLMs. Existing scientific VLM benchmarks cover broad scientific question answering, microscopy image understanding\cite{b6,b7,b8}, chart reasoning\cite{b9,b10}, document understanding\cite{b11,b12}, and medical image analysis. However, existing scientific and materials-science VLM datasets still have limitations in research focus, data construction, and evaluation design: 

\textbf{(1) Lack of evaluation for complex scientific images.} Current datasets focus on simple or common scientific images, leaving complex domain-specific diagrams, such as phase diagrams, insufficiently evaluated.  

\textbf{(2) Incomplete exploitation of original data.} Image–text pairs are often constructed using only captions or local paragraphs, without fully leveraging the context and information present in the original literature.  

\textbf{(3) Limited reliability of textual quality.} Texts in the datasets are selectively written to address specific research questions, making single-ground-truth-based evaluations potentially misaligned and less trustworthy.

\vspace{1em}
\textbf{\textit{How can we design benchmarks that reliably evaluate VLMs on complex, domain-specific scientific images, exemplified by phase diagrams?}}
\vspace{1em}

To address these three limitations, we propose \textbf{MatPhaseBench}, a vision-language benchmark specifically designed for materials phase diagram understanding. It focuses on semantically dense scientific images represented by phase diagrams and provides deeper, more fine-grained evaluation from three aspects: \textbf{complex scientific image understanding}, \textbf{comprehensive image--text matching}, and \textbf{high-quality human-supervised text acquisition}.

Based on MatPhaseBench, we evaluate 13 representative VLMs, including both closed-source and open-source models. The experimental results show that current VLMs still struggle substantially with phase diagram understanding. Even the best-performing model achieves only 0.407 in BERTScore Recall. These results indicate that existing VLMs have difficulty covering key scientific information and reproducing expert-level organization in phase diagram descriptions. The comparison also shows that closed-source models generally perform better, but several open-source or smaller models remain competitive in specific metrics, suggesting that phase diagram understanding cannot be solved by language generation ability or parameter scale alone.

Our main contributions are as follows:
\begin{itemize}
    \item \textbf{Targeting complex scientific image understanding.} We construct \textbf{MatPhaseBench}, a high-quality and trustworthy benchmark for complex scientific image understanding in materials phase diagrams. It moves beyond simple objective tests and formulates phase diagram understanding as open-ended tasks requiring deep visual comprehension and materials-science reasoning.

    \item \textbf{Comprehensive image--text matching.} We propose a two-stage image--text matching strategy for constructing phase diagram datasets, consisting of direct matching and associative matching. This strategy preserves both directly aligned descriptions and contextually related textual evidence, enabling the construction of semantically richer and more reliable phase diagram--text samples.

    \item \textbf{High-quality human-supervised text acquisition.} We manually review and validate all descriptions to ensure their relevance, completeness, and reliability. This strict human-supervised acquisition process provides high-quality textual supervision for evaluating VLMs on materials phase diagram understanding.
\end{itemize}

\section{Related Works}
\subsection{Vision-Language Models}

In recent years, vision-language models (VLMs) have rapidly evolved from early systems for image recognition, captioning, and image--text matching into native multimodal models capable of supporting complex workflows. Current closed-source models, exemplified by GPT-5.5~\cite{b13,b14,b15}, exhibit strong capabilities in long-context multimodal input, document parsing, chart and PDF reasoning, tool use, and agentic workflows, performing competitively on scientific and reasoning-intensive benchmarks such as MMMU(Pro)~\cite{b16,b17}, GPQA Diamond~\cite{b18}, FrontierMath~\cite{b19}, GeneBench~\cite{b20}, and SciCode~\cite{b21}. Similarly, open-source models, represented by Qwen3.6~\cite{b22,b23,b24}, have advanced in multimodal fusion, long-document and multi-page PDF understanding, GUI and video reasoning, spatial relationship modeling, visual grounding, and tool-assisted workflows.

Overall, the evolution of VLMs indicates a clear shift from perception-oriented visual understanding toward reasoning-intensive, tool-augmented, and domain-adaptive multimodal intelligence. This shift is particularly important for scientific images, which encode experimental evidence, theoretical assumptions, and specialized reasoning. Materials phase diagrams exemplify such complexity, requiring the interpretation of axes, composition ranges, phase regions, boundary curves, invariant reactions, and thermodynamic implications. Existing general-purpose VLM benchmarks cannot determine whether models truly understand these concepts or merely perform OCR or chart recognition, motivating a dedicated benchmark for phase diagram understanding.

\subsection{VLM Benchmark in the Scientific Domain}

Scientific-domain VLM benchmarks are designed to evaluate multimodal understanding under knowledge-intensive settings. Compared with general visual question answering or captioning benchmarks, scientific benchmarks typically require models to combine explicit visual structures with domain knowledge. Materials-science benchmarks remain scarce; here, we also include representative datasets in this domain for comparison.

\begin{table*}[t]
\centering
\caption{Comparison of Vision-Text Multimodal Benchmarks}
\label{tab:benchmark_comparison}
\renewcommand{\arraystretch}{1.25}
\footnotesize
\setlength{\tabcolsep}{5pt}

\begin{tabular}{
>{\raggedright\arraybackslash}p{2.8cm}
>{\centering\arraybackslash}p{2.5cm}
>{\centering\arraybackslash}p{2.4cm}
>{\centering\arraybackslash}p{3.4cm}
>{\centering\arraybackslash}p{3.0cm}
>{\centering\arraybackslash}p{2.4cm}
}
\toprule
\textbf{Benchmark} 
& \textbf{Domain} 
& \textbf{Publication-derived} 
& \textbf{Comprehensive Information Matching} 
& \textbf{Exclusively Human-supervised}
& \textbf{Open-ended Task} \\
\midrule

\textbf{MatPhaseBench (Ours)}
& Materials Science 
& \checkmark 
& \checkmark\checkmark 
& \checkmark 
& \checkmark \\

MATRIX\cite{b25}
& Materials Science 
& \checkmark 
& \checkmark 
& $\times$ 
& \checkmark \\

MicroscopyGPT\cite{b26}
& Materials Science 
& $\times$ 
& $\times$ 
& $\times$ 
& $\times$ \\

SEM-VLM Dataset\cite{b27}
& Materials Science 
& \checkmark 
& \checkmark 
& $\times$ 
& $\times$ \\

Cephalo\cite{b28}
& Materials Science 
& \checkmark 
& $\times$ 
& $\times$ 
& \checkmark \\

MicroVQA\cite{b8}
& Biology / Biomedical 
& $\times$ 
& -- 
& \checkmark 
& $\times$ \\

MMSci\cite{b29} 
& Science 
& \checkmark 
& $\times$ 
& $\times$ 
& $\times$ \\

MAC\cite{b30} 
& Science 
& \checkmark 
& $\times$ 
& \checkmark 
& $\times$ \\

Micro-Bench\cite{b6} 
& Biology / Biomedical 
& \checkmark 
& $\times$ 
& \checkmark 
& \checkmark \\

MME-SCI\cite{b31} 
& Multi-discipline 
& \checkmark 
& $\times$ 
& \checkmark 
& $\times$ \\

ProJudgeBench\cite{b32} 
& Multi-discipline 
& \checkmark 
& $\times$ 
& \checkmark 
& $\times$ \\

mmJEE-Eval\cite{b33} 
& Multi-discipline 
& \checkmark 
& $\times$ 
& \checkmark 
& $\times$ \\

MDK12-Bench\cite{b34} 
& Multi-discipline 
& $\times$ 
& -- 
& $\times$ 
& $\times$ \\

MMMU\cite{b35} 
& Multi-discipline 
& $\times$ 
& -- 
& $\times$ 
& $\times$ \\

\bottomrule
\end{tabular}

\vspace{0.3em}
\begin{minipage}{0.95\textwidth}
\footnotesize
$^{\mathrm{a}}$Two \checkmark{} symbols indicate that the data went through two matching stages.
\end{minipage}

\end{table*}

Existing scientific VLM benchmarks have evolved from exam-style assessments to research-oriented evaluation. Early benchmarks such as MMMU, MDK12-Bench, mmJEE-Eval, and MME-SCI focus on broad multimodal reasoning across educational tasks, while more recent efforts leverage real research data: MAC uses journal cover images, Micro-Bench and Verma et al. focus on biological microscopy, and MicroVQA and MMSci draw from scientific literature and experimental reports. In materials science, datasets such as SEM-VLM, MicroscopyGPT, MATRIX, and Cephalo target microstructures, characterization images, and literature. Despite these foundations, existing benchmarks are often limited in research-level difficulty, fine-grained scientific annotation, multi-dimensional evaluation, and suitability for complex materials phase diagram understanding. 

As shown in Table~\ref{tab:benchmark_comparison}, many are not publication-derived, rely only on captions or local paragraph matching, lack fully human-supervised quality control, or are formulated as closed-form tasks. MatPhaseBench addresses these limitations by constructing a publication-derived benchmark from classical phase diagram literature, adopting a two-stage image--text matching strategy to capture both directly aligned and contextually associated descriptions, applying fine-grained human validation to all textual samples, and introducing 5 semantic dimensions to comprehensively evaluate VLM understanding of phase diagrams. In this way, MatPhaseBench provides a high-quality, rustworthy, open-ended benchmark for assessing complex scientific image understanding in materials science.

\section{MatPhaseBench Benchmark}

\begin{figure*}[htbp]
\centering
\includegraphics[
    width=0.95\textwidth,
    trim=0cm 3cm 0cm 3cm,
    clip
]{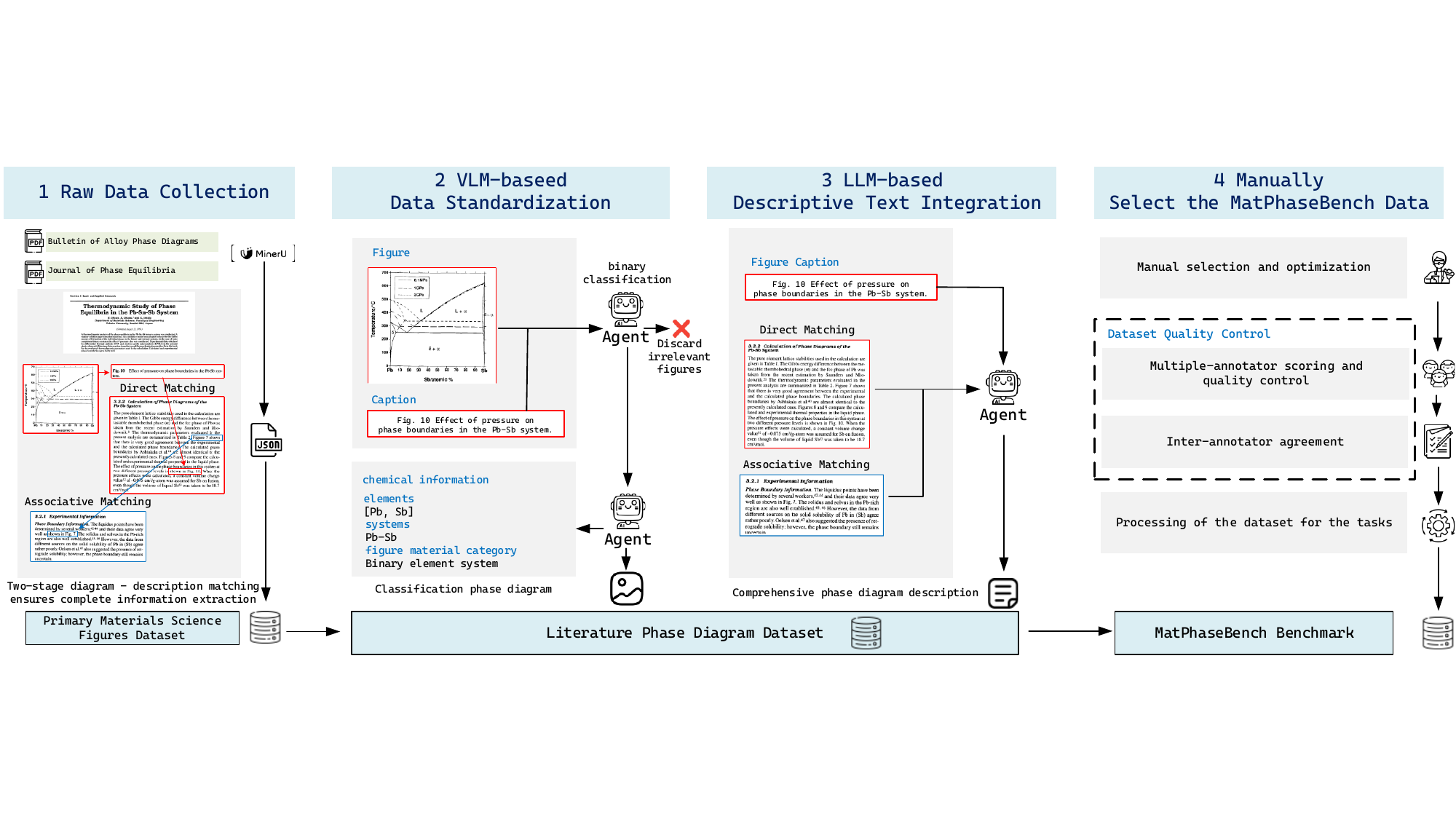}
\caption{Data processing pipeline.}
\label{fig:phase_pipeline}
\end{figure*}

In this section, we describe the construction of MatPhaseBench in detail, including data collection, the processing pipeline, quality control, and semantic-dimension annotation. MatPhaseBench is designed to evaluate whether VLMs can understand materials phase diagrams at a scientific semantic level, rather than only recognizing visual patterns or generating generic captions.

\subsection{Task Formulation}

In MatPhaseBench, each sample consists of a phase diagram image $I_i$ and its corresponding textual description $T_i^{*}$ extracted from materials science literature. The task requires a vision-language model $f_\theta$ to generate a free-form textual description $\hat{T}_i$ that accurately captures the key information in the phase diagram and is semantically consistent with the literature description. Formally, the generation process can be expressed as

\begin{equation}
\hat{T}_i = \arg\max_T P_\theta(T \mid I_i),
\end{equation}

where $I_i$ denotes the input image, $\hat{T}_i$ the model-generated description, and $P_\theta(T \mid I_i)$ the probability assigned by the model. The generated description is evaluated based on its textual and semantic alignment with the literature reference using automatic metrics, specifically BERTScore, ROUGE-1, and ROUGE-L. Let $m$ denote an evaluation metric. The average score over the entire dataset is defined as

\begin{equation}
S_m = \frac{1}{N} \sum_{i=1}^{N} m(\hat{T}_i, T_i^{*}),
\end{equation}

where $N$ is the number of samples in the benchmark and $m \in \{\mathrm{BERTScore}, \mathrm{ROUGE\text{-}1}, \mathrm{ROUGE\text{-}L}\}$.

\subsection{Overview of MatPhaseBench}

Although VLMs have advanced rapidly in general multimodal tasks, materials phase diagrams remain a demanding test of scientific image understanding because they require models to connect visual topology—axes, curves, regions, and labels—with materials-science semantics such as phase stability, thermodynamic constraints, and invariant reactions. To address this challenge, we propose MatPhaseBench, a high-quality, trustworthy, and open-ended benchmark for evaluating VLM understanding of materials phase diagrams. MatPhaseBench is constructed from classic phase-diagram literature published between 1980 and 2003 in \textit{Bulletin of Alloy Phase Diagrams}\cite{b36} and \textit{Journal of Phase Equilibria}\cite{b36}. As shown in Table~\ref{tab:phase_datasets}, starting from 3,681 PDF papers, we build an initial multimodal dataset covering 1,949 materials systems and 4,763 phase diagram--text matched samples. From these, 200 high-quality samples are manually selected and reviewed by three PhD students engaged in interdisciplinary computational materials research to form the final evaluation set.

\begin{table}[htbp]
\centering
\caption{Comparison of Phase Diagram Datasets}
\label{tab:phase_datasets}
\scriptsize
\renewcommand{\arraystretch}{1.25}
\setlength{\tabcolsep}{3pt}

\begin{tabular}{
>{\centering\arraybackslash}m{0.32\columnwidth}
>{\centering\arraybackslash}m{0.12\columnwidth}
>{\centering\arraybackslash}m{0.13\columnwidth}
>{\centering\arraybackslash}m{0.16\columnwidth}
>{\centering\arraybackslash}m{0.12\columnwidth}
}
\toprule
\textbf{Dataset} 
& \textbf{Papers} 
& \textbf{Phase Diagrams} 
& \textbf{Material Systems} 
& \textbf{Elements} \\
\midrule
Literature phase diagram dataset 
& 3681 
& 4763 
& 1949 
& 90 \\

MatPhaseBench 
& 200 
& 200 
& 189 
& 70 \\
\bottomrule
\end{tabular}
\end{table}

\subsection{ Dataset Construction}
As shown in Fig.~\ref{fig:phase_pipeline}, the construction pipeline of MatPhaseBench consists of 5 stages: Raw Data Collection, Phase Diagram Data Standardization, Phase Diagram Description Integration, Quality Control, and Task-specific Processing. The detailed procedures for each stage are described in this section.

\textbf{Raw Data Collection.}

We collect PDF papers from classical materials journal, and use MinerU\cite{b37} for structured document parsing to extract figures, captions, and paragraph text. Based on the parsed results, we construct image--text pairs through a two-stage matching process. In the \textbf{\textit{direct matching stage}}, each phase diagram is matched with its corresponding caption or paragraphs that explicitly refer to the target figure. In the \textbf{\textit{associative matching stage}}, when these explicitly referring paragraphs also discuss other figures, the textual descriptions associated with those related figures are further matched to the target phase diagram as contextual evidence. This process enables us to build an initial multimodal dataset in which each phase diagram is linked to both directly matched descriptions and contextually related textual information.

\textbf{Phase Diagram Data Standardization.}

We use Qwen3.6-Plus\cite{b22} to process the initial multimodal dataset. First, a binary classification step is applied to screen whether a figure is a phase diagram. Then, materials systems are semantically parsed to extract chemical elements and categorize the system. This process maps general materials images from the literature into a structured subset of phase diagram samples, enabling subsequent semantic annotation and evaluation.

\textbf{Phase Diagram Description Integration.}

Because textual descriptions of a phase diagram are often scattered across paragraphs, experimental discussions, and thermodynamic analyses, we employ an LLM-assisted (GLM-5.1\cite{b38}) process to clean and reconstruct the raw descriptions. This approach retains only sentences directly relevant to the target diagram, its materials system, phase evolution, thermodynamic behavior, or essential cross-figure comparisons, while preserving original academic phrasing and traceability and enhancing consistency, semantic focus, and image–text alignment without introducing unsupported inferences.

\subsection{Dataset Quality Control}

To construct the final benchmark, we manually review and select 200 high-quality samples from the initial dataset. These samples cover 189 distinct material systems and include both multi-panel and single-image phase diagrams. Each sample is evaluated from three perspectives: \textbf{\textit{completeness}}, \textbf{\textit{accuracy}}, and \textbf{\textit{factuality}}. The detailed annotation guidelines are presented in APPENDIX Table~\ref{tab:annotation_guidelines}. Each dimension is scored using a discrete scale of (3, 2, 1, 0). Through this multi-dimensional human evaluation mechanism, we systematically control the quality of phase diagram descriptions from three perspectives: information coverage, image–text alignment, and evidential faithfulness.

Prior to formal annotation, three rounds of pilot annotation were conducted, with the guidelines revised after each round. To assess reliability, the first 10\% of samples, namely 20 samples, were evaluated for inter-annotator agreement using Simple Agreement, Gwet's AC1\cite{b39}, and Cohen's $\kappa$\cite{b40}. As shown in Table~\ref{tab:IAA}, the results indicate high consistency across completeness, accuracy, and factuality, with Simple Agreement and Gwet's AC1 above 0.92, while the lower $\kappa$ values mainly reflect the concentration of high-quality samples rather than substantive disagreement. These results suggest that the annotators had reached a stable scoring consensus. The remaining samples were then annotated following the finalized guidelines, and all disputed or low-quality samples were manually reviewed and corrected to ensure annotation reliability.

\begin{table}[htbp]
\centering
\caption{IAA of the annotation process}
\label{tab:IAA}
\renewcommand{\arraystretch}{1.15}
\footnotesize
\setlength{\tabcolsep}{6pt} % 调整列间距

\begin{tabular}{lccc}
\toprule
\textbf{Metric Type} & \textbf{Simple Agreement} & \textbf{Gwet's AC1} & \textbf{Cohen's $\kappa$} \\
\midrule
Completeness & 0.970 & 0.970 & 0.376 \\
Accuracy     & 0.924 & 0.920 & 0.578 \\
Factuality   & 0.941 & 0.938 & 0.540 \\
\bottomrule
\end{tabular}
\end{table}

\subsection{Multi-dimensional Semantic Categorization}

To improve the targetedness and interpretability of evaluation, MatPhaseBench does not treat phase diagram description as an unrestricted free-form captioning task alone. Instead, we summarize the semantic content of phase diagram descriptions into 5 predefined semantic dimensions, as shown in Table~\ref{tab:semantic_dimensions}. During dimension annotation, we assign semantic dimensions according to information explicitly stated in the ground-truth descriptions, rather than implicit information that would require additional inference. It should also be noted that although the types of semantic dimensions are predefined, not every sample covers all 5 dimensions. Statistics show that each sample contains four semantic dimensions on average, indicating that phase diagram understanding is inherently multidimensional.

\begin{table}[htbp]
\centering
\caption{Semantic Dimensions of MatPhaseBench}
\label{tab:semantic_dimensions}
\scriptsize
\renewcommand{\arraystretch}{1.3}
\setlength{\tabcolsep}{4pt}

\begin{tabular}{
>{\centering\arraybackslash}m{0.22\columnwidth}
>{\centering\arraybackslash}m{0.58\columnwidth}
>{\centering\arraybackslash}m{0.15\columnwidth}
}
\toprule
\textbf{Semantic Dimension} & \textbf{Content Description} & \textbf{Number of Samples} \\
\midrule

Materials system
& The materials system involved in the phase diagram, such as binary, ternary, or multicomponent systems.
& 199 \\

Phase diagram type
& The type of phase diagram, such as an isothermal section, vertical section, liquidus projection, or composite diagram.
& 121 \\

Phase-diagram coverage
& Whether the phase diagram is complete or partial, and the composition range covered by the diagram.
& 73 \\

Phase region and phase boundary recognition
& Recognition of single-phase, two-phase, and three-phase regions, as well as the corresponding phase boundaries.
& 185 \\

Invariant reactions
& Recognition of invariant reactions such as congruent melting, eutectic, eutectoid, monotectic, metatectic, peritectic, and syntectic reactions, including the associated temperatures and compositions.
& 116 \\

\bottomrule
\end{tabular}
\end{table}

\section{Phase Diagram Understanding Task}

\begin{figure}[htbp]
\centering
\makebox[\columnwidth][c]{%
\includegraphics[
    width=1\columnwidth,
    trim=4cm 0cm 4cm 0cm,
    clip
]{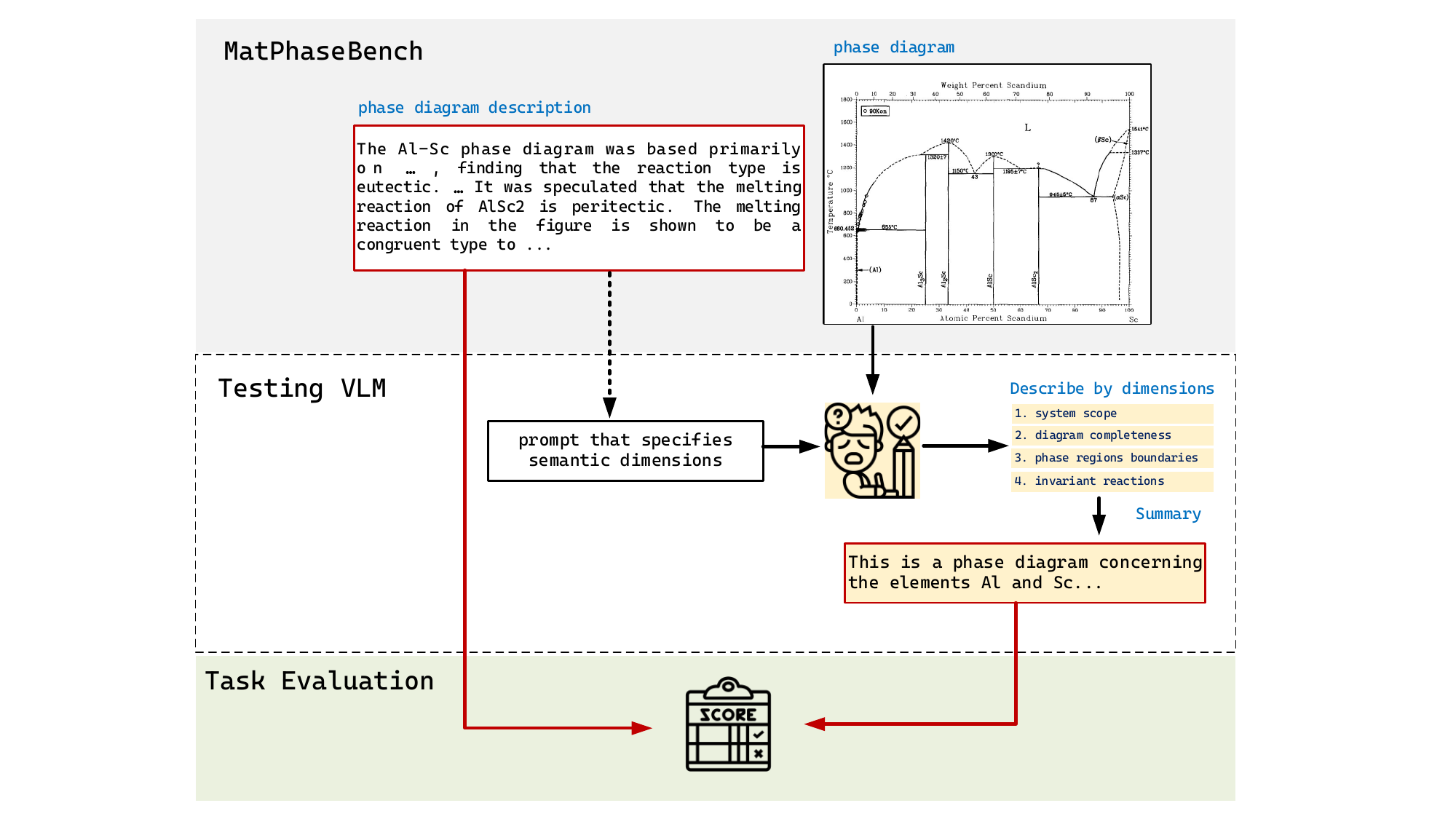}
}
\caption{Schematic diagram of phase diagram understanding task.}
\label{fig:phase_diagram_task}
\end{figure}

\subsection{Task Design}

MatPhaseBench defines a semantic-dimension-guided phase diagram understanding task in which a VLM generates scientific descriptions from predefined semantic perspectives. While the task is open-ended, the semantic dimensions guide the model to focus on information aligned with human experts' attention in the literature. This approach addresses the selective nature of phase diagram descriptions in scientific papers, where experts may emphasize specific phase regions, reaction points, comparisons with prior assessments, or thermodynamic implications. Because a single Ground Truth cannot cover all reasonable descriptions, free-form caption evaluation may misrepresent valid outputs. To mitigate this, as shown in Fig.~\ref{fig:phase_diagram_task}, each test sample is assigned multi-label semantic dimension annotations embedded into the prompt, reducing narrative-viewpoint bias, improving comparability with literature-derived references, and supporting dimension-aligned automatic evaluation and fine-grained error analysis.

\subsection{Evaluation Methodology}

For automatic evaluation, we use conventional text-generation metrics and semantic similarity metrics. Specifically, we report ROUGE-1\cite{b41}, ROUGE-L\cite{b41} and BERTScore\cite{b42} with Recall, F1 scores. ROUGE evaluates lexical overlap between the generated description and the Ground Truth, while BERTScore evaluates semantic similarity using contextual embeddings. Since phase diagram descriptions may use different wording to express similar scientific meanings, semantic metrics are important complements to lexical metrics.

The goal of MatPhaseBench evaluation is not to require a model to reproduce the Ground Truth word by word. Instead, the benchmark evaluates whether the model covers the key semantic information emphasized in the expert description. For this reason, Recall is particularly important: it measures how much of the Ground Truth information is captured by the model output. F1 provides a complementary balance between coverage and precision. 

\section{Experiments}

\subsection{Experimental Setup}

We evaluate 13 state-of-the-art VLMs on MatPhaseBench as baselines, including both closed-source models (GPT-5\cite{b13}, Gemini 3 series\cite{b15}, Claude Opus 4.8\cite{b14}, two Qwen3.6\cite{b22} variants, GLM-5V-Turbo\cite{b43}) and open-source models (Qwen3.6 open version, GLM-4.6V\cite{b38} series, Llama-4\cite{b24} series). Prompts are constructed in a zero-shot manner, dynamically incorporating pre-assigned semantic dimensions to guide the models’ interpretation of each phase diagram. All models are run in inference mode with standardized parameters: temperature = 0, top-p = 0.8, top-k = 20. 

\subsection{Main Findings}

This section examines representative low-scoring samples from Claude Opus 4.8 on the phase diagram understanding task, as shown in Table~\ref{tab:key_finding}, with particular emphasis on cases that provide critical insights. Here, “human expert” refers to the ground-truth descriptions in MatPhaseBench. Fine-grained comparisons show that low scores reflect not only model limitations, but also inherent challenges in phase diagram understanding: the gap between visual recognition and materials reasoning, the misalignment between model outputs and expert domain awareness, the limited comparative reasoning ability for composite phase diagrams, and the limitations of conventional metrics in evaluating complex scientific image understanding.

\begin{table*}[htbp]
\centering
\caption{Key Findings from Claude's Phase Diagram Understanding}
\label{tab:key_finding}

\renewcommand{\arraystretch}{1.18}
\setlength{\tabcolsep}{3pt}

\begin{tabular}{
>{\centering\arraybackslash}m{0.05\textwidth}
>{\centering\arraybackslash}m{0.26\textwidth}
>{\centering\arraybackslash}m{0.22\textwidth}
>{\centering\arraybackslash}m{0.22\textwidth}
>{\centering\arraybackslash}m{0.16\textwidth}
}
\toprule
\textbf{Case} &
\textbf{Phase Diagram} &
\textbf{Claude's Understanding} &
\textbf{Ground Truth} &
\textbf{Findings} \\
\midrule

Case 1 &
\includegraphics[width=0.24\textwidth]{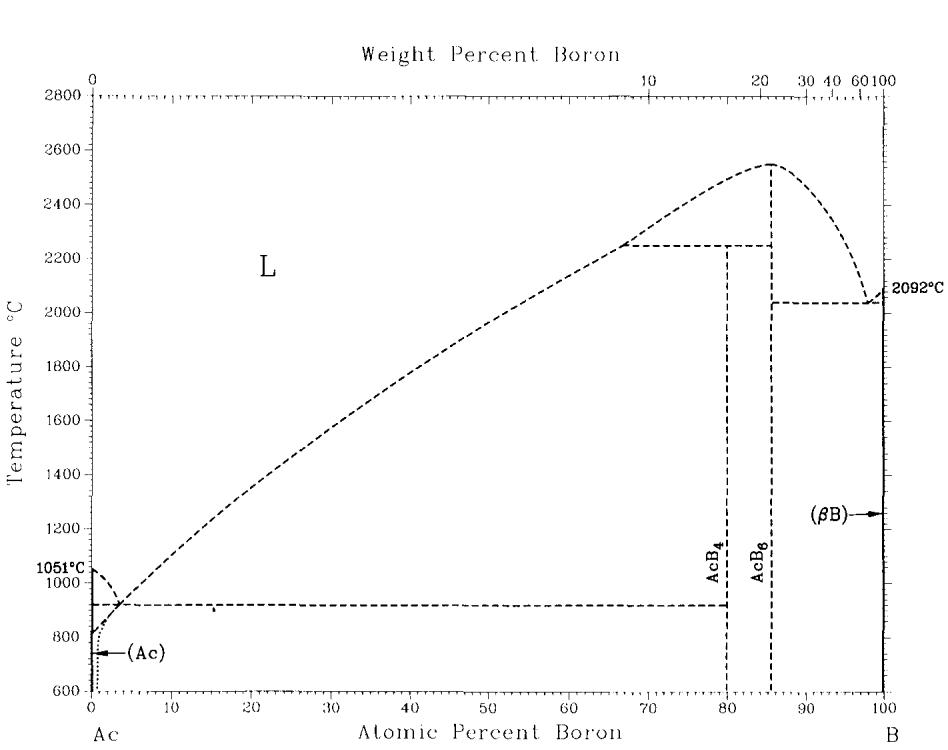} &
This is a complete binary Ac-B (actinium-boron) temperature-composition phase diagram spanning from pure Ac to pure B. The vertical axis shows Temperature in $^\circ$C from 600 to 2800 $^\circ$C, the lower horizontal axis is Atomic Percent Boron (0--100), and the upper horizontal axis is the corresponding Weight Percent Boron, illustrating the nonlinear scale relationship. ... &
... Here we recall that the extrapolated liquidus cannot cross the 0 at.\% solute line, because the slope of the Gibbs energy function of the liquid phase always (except at 0 K) has a negative infinity value at 0 at.\% due to the $RT X \log X$ term, which derives from the contribution of the ideal entropy of mixing. ... &
VLMs usually focus on surface-level and visually explicit information, whereas human experts pay more attention to deeper materials-science theoretical analysis. \\
\midrule

Case 2 &
\includegraphics[width=0.24\textwidth]{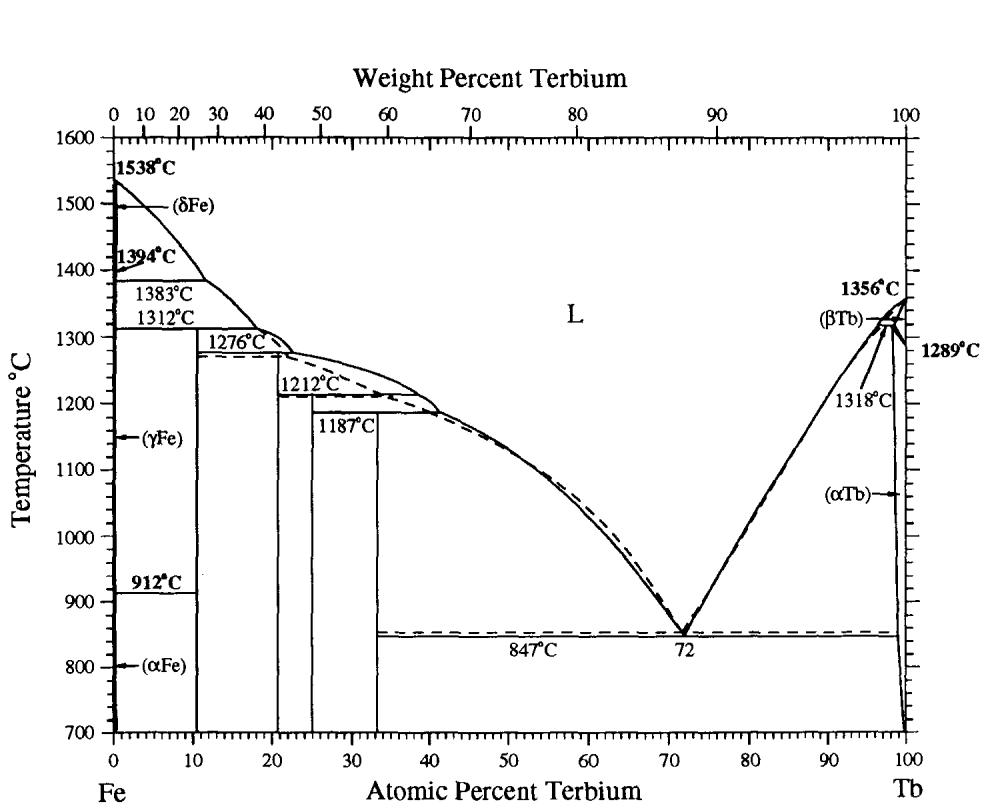} &
The liquidus descends from the Fe melting point through a series of horizontal invariant isotherms at 1383, 1312, 1276, 1212, and 1187 $^\circ$C, which bound the intermediate phases and are consistent with peritectic (L + solid $\rightarrow$ intermediate) and eutectic (L $\rightarrow$ solid + solid) reactions. ... Dashed curves indicate uncertain or alternative liquidus and lower-boundary positions ... &
... The slope in the diagram shown by solid lines is much steeper causing the liquidus composition in the peritectic reaction of $\mathrm{Fe}_{17}\mathrm{Tb}_2$ to be too Fe rich ... &
VLMs usually provide general, global, and surface-level descriptions, whereas human experts focus more on specific domain research questions. \\
\midrule

Case 3 &
\includegraphics[width=0.24\textwidth]{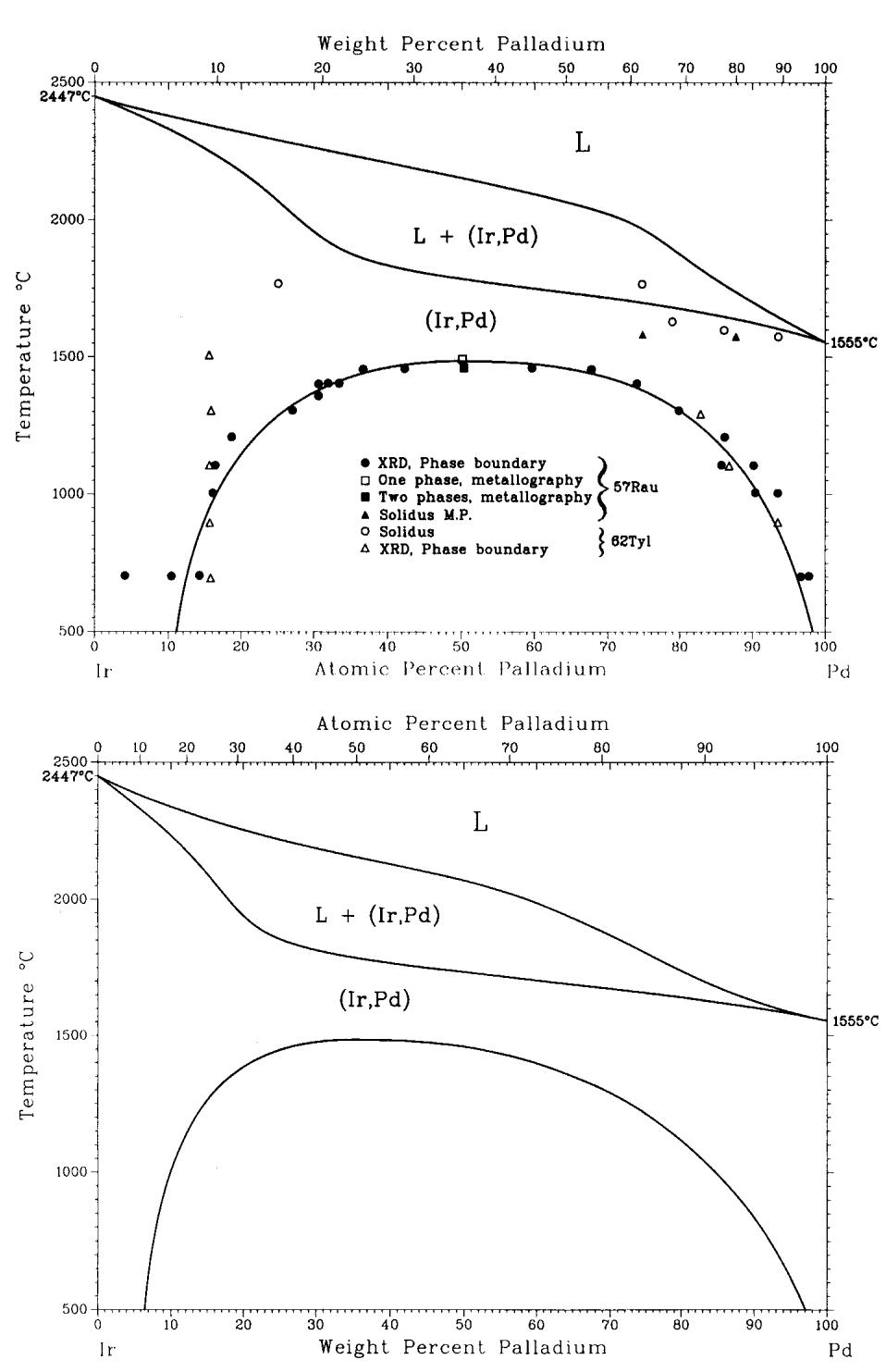} &
The figure presents the binary Ir-Pd (iridium-palladium) system as a combined pair of complete temperature-composition phase diagrams. ... Both span the full composition range from pure Ir (melting point 2447~$^\circ$C) to pure Pd (melting point 1555~$^\circ$C) and the full temperature window, so no subsystem or element-rich region is omitted ... &
In the absence of experimental data over the entire composition range for the liquidus and the solidus phase boundaries, the liquidus and the solidus are evaluated, assuming a regular solution behavior and the model. ... The calculated solidus temperatures are higher than the results of both. ... &
VLMs show insufficient capability in comparative analysis of composite phase diagrams. \\
\midrule

Case 4 &
\includegraphics[width=0.24\textwidth]{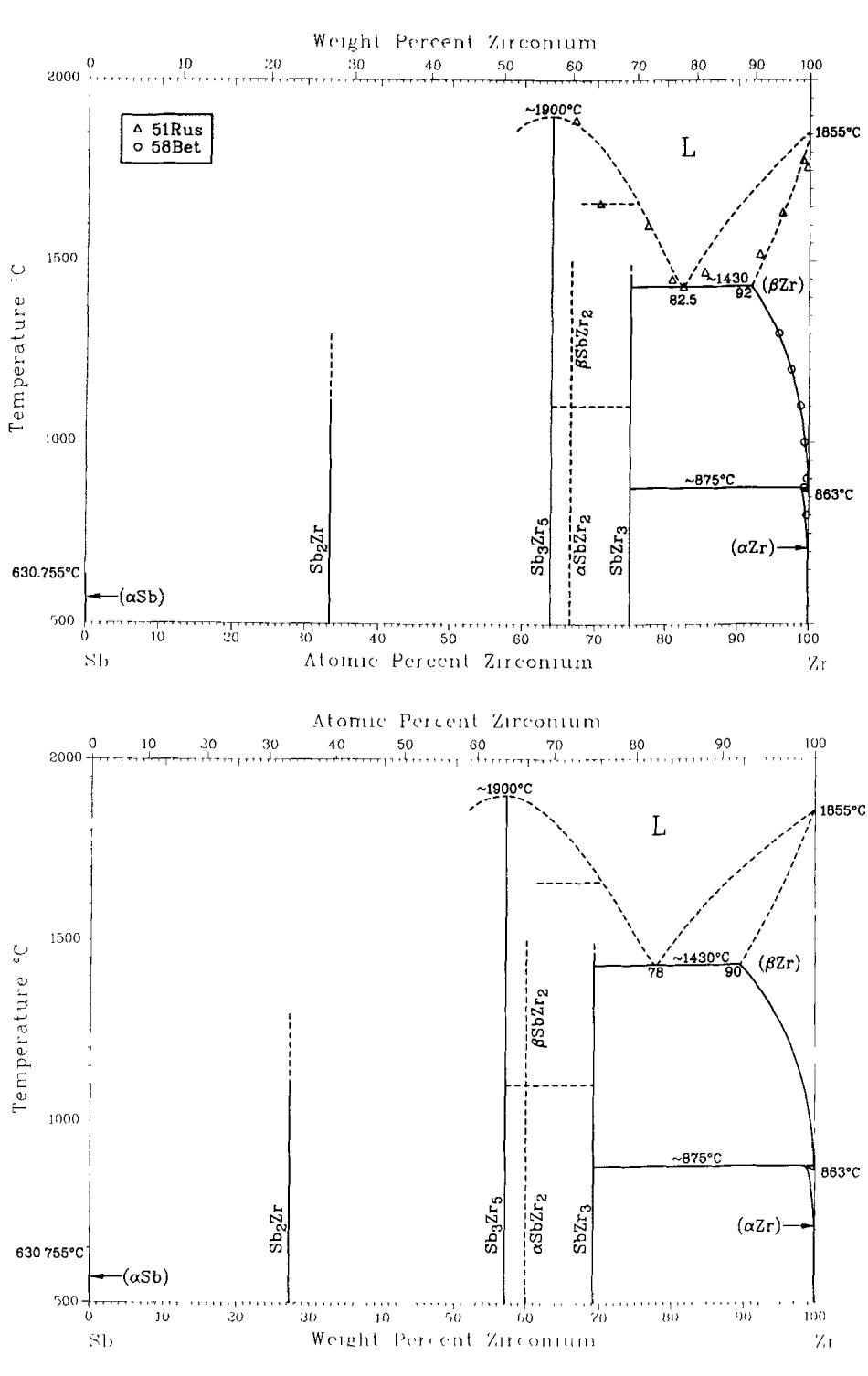} &
... Key invariant features include a congruent melting maximum at $\sim$1900~$^\circ$C (L $\rightarrow$ Sb$_3$Zr$_5$); a eutectic at $\sim$1430~$^\circ$C on the Zr-rich side (L $\rightarrow$ SbZr$_3$ + ($\beta$Zr)), with composition markers near 82.5/92 at.\% Zr (upper) and 78/90 at.\% Zr (lower); a polymorphic transformation $\beta$SbZr$_2$ $\rightarrow$ $\alpha$SbZr$_2$ near $\sim$1100~$^\circ$C; and the ($\beta$Zr) $\rightarrow$ ($\alpha$Zr) transformation near $\sim$875~$^\circ$C ending at $\sim$863~$^\circ$C at the Zr edge. ... &
... Because the crystal structure of SbZr$_2$ is similar to that of Sb$_3$Zr$_5$ reported later, the melting point of SbZr$_2$ at $\sim$1900~$^\circ$C is regarded as that of Sb$_3$Zr$_5$ in the present evaluation. ... The composition of ($\beta$Zr) at the L $\leftrightarrow$ SbZr$_3$ + ($\beta$Zr) eutectic temperature (1430~$^\circ$C) is shown at 92 at.\% Zr, which was obtained by an extrapolation of a plot of log [solubility of Sb in ($\beta$Zr)] vs 1 / T. The peritectoid transformation temperature of ($\beta$Zr) $\leftrightarrow$ ($\alpha$Zr) was estimated to be 875~$^\circ$C from an impurity-Sb-Zr ternary phase diagram. ... &
The conventional evaluation metrics used in this work are inadequate for assessing phase diagram understanding with complex logical reasoning and need further improvement. \\
\bottomrule

\end{tabular}
\end{table*}

\begin{itemize}
    \item \textbf{VLMs Surface-Level Perception vs. Expert Deep Theoretical Reasoning} \\
    As shown in Table~\ref{tab:key_finding} Case 1, VLMs can often identify explicit information in a phase diagram, such as axes, phase labels, and boundary curves. However, human experts are more concerned with deeper theoretical issues in materials science that are not directly visible, such as thermodynamic consistency, the physical plausibility of phase boundaries, the construction process of assessed diagrams, and composition calculations derived from thermodynamic rules.

    \item \textbf{VLMs Lack Domain Problem Insight and Analytical Experience} \\
    As shown in Table~\ref{tab:key_finding} Case 2, in phase diagram understanding, VLMs lack the problem-awareness and analytical experience that human experts demonstrate in materials science. Whereas experts focus on identifying and analyzing critical domain-specific issues—such as the reliability of a boundary, or comparison with experimental data—VLM outputs tend to provide broad interpretations of the entire diagram, often overlooking the targeted reasoning that guides expert analyses. This gap reflects both a limitation in model understanding and the challenge of aligning task objectives and evaluation with domain-focused scientific reasoning.

    \item \textbf{VLMs Lack Composite Phase Diagram Understanding} \\
    As shown in Table~\ref{tab:key_finding} Case 3, some samples contain multiple versions of a diagram for the same materials system. In such cases, VLMs tend to summarize the shared structure of the diagrams, but they often fail to identify differences among versions. Expert descriptions, by contrast, often emphasize the basis and significance of these differences. This indicates that VLMs still lack sufficient understanding and analytical capability for composite phase diagrams.

    \item \textbf{Limitations of Conventional Text Metrics in Evaluating Complex Scientific Images} \\
    As shown in Table~\ref{tab:key_finding} Case 4, automatic metrics also exhibit limitations in this task. Some model outputs contain valid scientific observations, but they receive low ROUGE or BERTScore values because their organization, terminology, granularity, or narrative order differs from the Ground Truth. Therefore, in complex open-ended scientific image understanding tasks, developing evaluation methods that better approximate human expert judgment and capture domain-specific reasoning remains an important direction for future research.
\end{itemize}

\subsection{Phase Diagram Understanding Results}

\begin{table*}[htbp]
\centering
\caption{Performance on MatPhaseBench. Bold indicates the best results, and underline indicates the second-best results.}
\label{tab:performance}
\scriptsize
\renewcommand{\arraystretch}{1.15}
\begin{tabular}{lcccccc}
\toprule
\multirow{2}{*}{\textbf{Models}} 
& \multicolumn{2}{c}{\textbf{BERTScore}} 
& \multicolumn{2}{c}{\textbf{ROUGE-1}} 
& \multicolumn{2}{c}{\textbf{ROUGE-L}} \\
\cmidrule(lr){2-3} \cmidrule(lr){4-5} \cmidrule(lr){6-7}
& \textbf{Recall} & \textbf{F1} 
& \textbf{Recall} & \textbf{F1} 
& \textbf{Recall} & \textbf{F1} \\
\midrule
\multicolumn{7}{c}{\textit{Closed-source Models}} \\
\midrule
Claude-Opus-4.8 
& \textbf{0.407} & 0.291 
& \textbf{0.438} & 0.341 
& \underline{0.225} & 0.171 \\
GLM-5V-Turbo 
& \underline{0.398} & 0.334 
& 0.400 & 0.343 
& 0.211 & 0.178 \\
GPT-5.5 
& 0.394 & 0.360 
& \underline{0.430} & 0.334 
& \textbf{0.229} & 0.174 \\
Gemini-3.1-Pro 
& 0.392 & \underline{0.388} 
& 0.390 & \underline{0.347} 
& 0.208 & 0.182 \\
Qwen3.6-Plus 
& 0.388 & 0.363 
& 0.356 & \textbf{0.348} 
& 0.189 & \underline{0.183} \\
Qwen3.6-Flash 
& 0.379 & 0.363 
& 0.349 & 0.342 
& 0.185 & 0.180 \\
Gemini-3.1-Flash-Lite 
& 0.363 & \textbf{0.396} 
& 0.311 & 0.343 
& 0.167 & \textbf{0.184} \\
\midrule
\multicolumn{7}{c}{\textit{Open-source Models}} \\
\midrule
Qwen3.6-27B 
& \textbf{0.383} & 0.375 
& \underline{0.349} & \textbf{0.343} 
& \underline{0.184} & \underline{0.180} \\
Qwen3.6-35B-A3B 
& \underline{0.380} & 0.369 
& \textbf{0.350} & \underline{0.342} 
& \textbf{0.187} & \textbf{0.181} \\
GLM-4.6V 
& 0.367 & 0.335 
& 0.308 & 0.301 
& 0.166 & 0.161 \\
GLM-4.6V-Flash 
& 0.345 & 0.336 
& 0.247 & 0.289 
& 0.134 & 0.156 \\
Llama-4-Maverick 
& 0.304 & \textbf{0.383} 
& 0.235 & 0.277 
& 0.128 & 0.150 \\
Llama-4-Scout 
& 0.303 & \underline{0.381} 
& 0.239 & 0.279 
& 0.130 & 0.151 \\
\midrule
\textbf{Avg. Performance} 
& 0.370 & 0.360 
& 0.339 & 0.325 
& 0.180 & 0.172 \\
\bottomrule
\end{tabular}

\vspace{2pt}
\parbox{0.98\textwidth}{
\scriptsize
}
\end{table*}

\indent\textbf{MatPhaseBench poses significant challenges to current VLMs.}

\indent The results in Table~\ref{tab:performance} show that phase diagram description remains highly challenging for current VLMs. Overall scores are low across all models. Even the best-performing model, Claude Opus 4.8, achieves only 0.407 in BERTScore Recall. ROUGE-L scores are also low, indicating that model outputs differ substantially from expert descriptions in long-range discourse structure, domain-specific narrative organization, and fine-grained materials-science semantics. These results suggest that current VLMs still have difficulty covering the key information contained in expert phase diagram descriptions.

\indent\textbf{Recall and F1 exhibit an inverted performance pattern.}

\indent In MatPhaseBench, the Ground Truth descriptions are derived from expert literature and are not exhaustive descriptions of all visible elements in a diagram. They usually focus on information considered important by materials experts. Therefore, Recall is especially relevant because it measures how much expert-emphasized information is captured by the model. In Table~\ref{tab:performance}, Claude Opus 4.8 achieves the best BERTScore Recall and ROUGE-1 Recall, with scores of 0.407 and 0.438, respectively, but it is not the best model in F1. In contrast, Gemini 3.1 Flash Lite achieves the best BERTScore F1 and ROUGE-L F1, but its Recall scores are lower. This indicates that some models generate longer and more comprehensive descriptions, increasing Recall while also introducing additional content that lowers F1. Other models generate more concise outputs that are locally closer to the Ground Truth, leading to higher F1 but weaker coverage.

\indent\textbf{Closed-source models show an overall advantage, while open-source and smaller models remain competitive.}

\indent The comparison between closed-source and open-source models shows that closed-source models generally have an advantage, but open-source and smaller models remain competitive in specific metrics. For example, Qwen3.6-Plus outperforms several larger or more general-purpose models, while Qwen3.6-27B performs better than Llama-4-Maverick on some metrics. These results suggest that phase diagram understanding cannot be explained by parameter scale alone. Model architecture, multimodal alignment, domain knowledge, and instruction-following behavior all influence performance.

Overall, the analyses in Sections B and C show that the low scores on MatPhaseBench should not be interpreted merely as failures of visual recognition. Instead, they reflect the high degree of professionalism, openness, and multidimensional semantics involved in phase diagram understanding. The main gaps between current VLMs and expert-level interpretation lie in three aspects: VLMs remain limited to surface-level perception rather than expert deep theoretical reasoning; they lack domain problem insight and analytical experience; and they still show insufficient understanding of composite phase diagrams. These findings also suggest that, for complex open-ended scientific image understanding tasks, developing evaluation methods that better approximate human expert judgment and capture domain-specific reasoning remains an important direction for future research.

\section{Conclusion}

This work introduces MatPhaseBench, a benchmark for evaluating vision-language models on materials phase diagram understanding. Built from classic phase-equilibrium literature, MatPhaseBench provides manually reviewed phase diagram--text samples with document-level semantic alignment, integrating evidence from captions, explicit figure-referenced paragraphs, and related discussions. It further organizes phase diagram understanding into five semantic dimensions: materials system, phase diagram type, phase-diagram coverage, phase region and phase boundary recognition, and invariant reactions.Experimental results on 13 representative VLMs show that current models remain far from expert-level understanding. Their outputs are largely limited to surface-level visual perception, with insufficient thermodynamic reasoning, limited materials-domain awareness, weak alignment with expert analytical focus, and poor discrimination of fine-grained differences in composite or multi-diagram settings.MatPhaseBench addresses these challenges by formulating phase diagram understanding as an open-ended scientific image understanding task, supported by comprehensive image--text matching and human-supervised text acquisition. By combining high-quality dataset construction, MatPhaseBench provides a reliable evaluation foundation for domain-specific VLMs, highlights the importance of systematically assessing complex scientific image understanding, and offers a pathway toward more trustworthy multimodal AI for materials-science discovery.

% \section*{References}

\appendix

\begin{table}[htbp]
\centering
\caption{Annotation Guidelines of MatPhaseBench}
\label{tab:annotation_guidelines}
\scriptsize
\renewcommand{\arraystretch}{1.25}
\begin{tabularx}{\columnwidth}{
>{\centering\arraybackslash}m{0.15\textwidth}
>{\centering\arraybackslash}X
}
\toprule
\textbf{Criterion} & \textbf{Annotation Dimension} \\
\midrule

\multirow{3}{*}{Completeness} & 
\parbox[c][2.5em][c]{\linewidth}{\centering Completeness of basic image information} \\
\cmidrule(lr){2-2}

& \parbox[c][2.5em][c]{\linewidth}{\centering Completeness of expert empirical description} \\
\cmidrule(lr){2-2}

& \parbox[c][2.5em][c]{\linewidth}{\centering Completeness of expert reasoning content} \\
\midrule

\multirow{3}{*}{Accuracy} & 
\parbox[c][2.5em][c]{\linewidth}{\centering Accuracy of image-related information filtering} \\
\cmidrule(lr){2-2}

& \parbox[c][2.5em][c]{\linewidth}{\centering Accuracy of readability-oriented rewriting} \\
\cmidrule(lr){2-2}

& \parbox[c][2.5em][c]{\linewidth}{\centering Accuracy of related supplementary information coverage} \\
\midrule

\multirow{3}{*}{Factuality} & 
\parbox[c][2.5em][c]{\linewidth}{\centering Factual consistency with the source text} \\
\cmidrule(lr){2-2}

& \parbox[c][2.5em][c]{\linewidth}{\centering Semantic faithfulness after simplified rewriting} \\
\cmidrule(lr){2-2}

& \parbox[c][2.5em][c]{\linewidth}{\centering Sentence-level evidence traceability} \\

\bottomrule
\end{tabularx}
\end{table}


\begin{thebibliography}{00}
\bibitem{b1} H. Okamoto, ``Al-Sc (aluminum-scandium),'' \textit{Journal of Phase Equilibria}, vol. 12, no. 5, pp. 612--613, 1991.
\bibitem{b2} J. Deng, W. Dong, R. Socher, L.-J. Li, K. Li, and F.-F. Li, ``Imagenet: A large-scale hierarchical image database,'' in \textit{2009 IEEE Conference on Computer Vision and Pattern Recognition}, pp. 248--255, 2009.
\bibitem{b3} A. Wang, A. Singh, J. Michael, F. Hill, O. Levy, and S. Bowman, ``GLUE: A multi-task benchmark and analysis platform for natural language understanding,'' in \textit{Proceedings of the 2018 EMNLP Workshop BlackboxNLP: Analyzing and interpreting neural networks for NLP}, pp. 353--355, 2018.
\bibitem{b4} S. Antol, A. Agrawal, J. Lu, M. Mitchell, D. Batra, C. L. Zitnick, and D. Parikh, ``Vqa: Visual question answering,'' in \textit{Proceedings of the IEEE International Conference on Computer Vision}, pp. 2425--2433, 2015.
\bibitem{b5} X. Yue, Y. Ni, K. Zhang, T. Zheng, R. Liu, G. Zhang, S. Stevens, D. Jiang, W. Ren, Y. Sun, et al., ``Mmmu: A massive multi-discipline multimodal understanding and reasoning benchmark for expert AGI,'' in \textit{Proceedings of the IEEE/CVF Conference on Computer Vision and Pattern Recognition}, pp. 9556--9567, 2024.
\bibitem{b6} A. Lozano, J. Nirschl, J. Burgess, S. R. Gupte, Y. Zhang, A. Unell, and S. Yeung-Levy, ``Micro-bench: A microscopy benchmark for vision-language understanding,'' \textit{Advances in Neural Information Processing Systems}, vol. 37, pp. 30670--30685, 2024.
\bibitem{b7} N. Gogoberidze and B. A. Cimini, ``Defining the boundaries: Challenges and advances in identifying cells in microscopy images,'' \textit{Current Opinion in Biotechnology}, vol. 85, p. 103055, 2024.
\bibitem{b8} J. Burgess, J. J. Nirschl, L. Bravo-Sánchez, A. Lozano, S. R. Gupte, J. G. Galaz-Montoya, Y. Zhang, Y. Su, D. Bhowmik, Z. Coman, et al., ``Microvqa: A multimodal reasoning benchmark for microscopy-based scientific research,'' in \textit{Proceedings of the IEEE/CVF Conference on Computer Vision and Pattern Recognition}, pp. 19552--19564, 2025.
\bibitem{b9} M. Huang, H. Lai, X. Zhang, W. Wu, J. Ma, L. Zhang, and J. Liu, ``Evochart: A benchmark and a self-training approach towards real-world chart understanding,'' in \textit{Proceedings of the AAAI Conference on Artificial Intelligence}, vol. 39, no. 4, pp. 3680--3688, 2025.
\bibitem{b10} K. Mukherjee, D. Ren, and D. Moritz, ``Encqa: Benchmarking vision-language models on visual encodings for charts,'' \textit{IEEE Transactions on Visualization and Computer Graphics}, 2025.
\bibitem{b11} M. S. Nacson, A. Aberdam, R. Ganz, E. B. Avraham, A. Golts, Y. Kittenplon, S. Mazor, and R. Litman, ``Docvlm: Make your VLM an efficient reader,'' in \textit{Proceedings of the Computer Vision and Pattern Recognition Conference}, pp. 29005--29015, 2025.
\bibitem{b12} H. Guo, X. Qin, J. Y. O. Yang, P. Zhang, G. Zeng, Y. Li, and H. Lin, ``Towards natural language-based document image retrieval: New dataset and benchmark,'' in \textit{Proceedings of the Computer Vision and Pattern Recognition Conference}, pp. 29722--29732, 2025.
\bibitem{b13} A. Singh, A. Fry, A. Perelman, A. Tart, A. Ganesh, A. El-Kishky, A. McLaughlin, A. Low, A. J. Ostrow, A. Ananthram, et al., ``OpenAI GPT-5 system card,'' \textit{arXiv preprint arXiv:2601.03267}, 2025.
\bibitem{b14} Anthropic, ``Claude Opus 4.8,'' 2025. [Online]. Available: \url{https://www.anthropic.com/news/claude-opus-4-8}. Accessed: Jun. 6, 2026.
\bibitem{b15} Google DeepMind, ``Gemini 3.1 Pro model card,'' 2025. [Online]. Available: \url{https://deepmind.google/models/model-cards/gemini-3-1-pro/}. Accessed: Jun. 6, 2026.
\bibitem{b16} X. Yue, Y. Ni, K. Zhang, T. Zheng, R. Liu, G. Zhang, S. Stevens, D. Jiang, W. Ren, Y. Sun, et al., ``Mmmu: A massive multi-discipline multimodal understanding and reasoning benchmark for expert AGI,'' in \textit{Proceedings of the IEEE/CVF Conference on Computer Vision and Pattern Recognition}, pp. 9556--9567, 2024.
\bibitem{b17} X. Yue, T. Zheng, Y. Ni, Y. Wang, K. Zhang, S. Tong, Y. Sun, B. Yu, G. Zhang, H. Sun, et al., ``Mmmu-pro: A more robust multi-discipline multimodal understanding benchmark,'' in \textit{Proceedings of the 63rd Annual Meeting of the Association for Computational Linguistics (Volume 1: Long Papers)}, pp. 15134--15186, 2025.
\bibitem{b18} D. Rein, B. L. Hou, A. C. Stickland, J. Petty, R. Y. Pang, J. Dirani, J. Michael, and S. R. Bowman, ``Gpqa: A graduate-level Google-proof Q\&A benchmark,'' \textit{arXiv preprint arXiv:2311.12022}, 2023.
\bibitem{b19} E. Glazer, E. Erdil, T. Besiroglu, D. Chicharro, E. Chen, A. Gunning, C. F. Olsson, J.-S. Denain, A. Ho, E. de Oliveira Santos, et al., ``FrontierMath: A benchmark for evaluating advanced mathematical reasoning in AI,'' \textit{arXiv preprint arXiv:2411.04872}, 2024.
\bibitem{b20} J. Li and A. Ho, ``GeneBench: Assessing AI agents for multi-stage inference problems in genomics and quantitative biology,'' \textit{bioRxiv}, pp. 2026--04, 2026.
\bibitem{b21} M. Tian, L. Gao, S. D. Zhang, X. Chen, C. Fan, X. Guo, R. Haas, P. Ji, K. Krongchon, Y. Li, et al., ``SciCode: A research coding benchmark curated by scientists,'' \textit{Advances in Neural Information Processing Systems}, vol. 37, pp. 30624--30650, 2024.
\bibitem{b22} ``Qwen3.6,'' GitHub. [Online]. Available: \url{https://github.com/QwenLM/Qwen3.6}.
\bibitem{b23} ``GLM-4.6V,'' Z AI Blog. [Online]. Available: \url{https://z.ai/blog/glm-4.6v}.
\bibitem{b24} ``LLaMA-4,'' Meta AI. [Online]. Available: \url{https://www.llama.com/docs/model-cards-and-prompt-formats/llama4/}.
\bibitem{b25} D. McGrath, C. Chong, R. Kulkarni, G. Ceder, and A. Kolluru, ``MATRIX: A Multimodal Benchmark and Post-Training Framework for Materials Science,'' \textit{arXiv preprint arXiv:2602.00376}, 2026.
\bibitem{b26} K. Choudhary, ``MicroscopyGPT: Generating atomic-structure captions from microscopy images of 2D materials with vision-language transformers,'' \textit{The Journal of Physical Chemistry Letters}, vol. 16, no. 27, pp. 7028--7035, 2025.
\bibitem{b27} Y. Cai and H. Wang, ``A visual language model enabling intelligent nanomaterial scanning electron micrograph annotation,'' \textit{Nanoscale}, vol. 17, no. 43, pp. 25136--25151, 2025.
\bibitem{b28} M. J. Buehler, ``Cephalo: Multi-Modal Vision-Language Models for Bio-Inspired Materials Analysis and Design,'' \textit{Advanced Functional Materials}, vol. 34, no. 49, p. 2409531, 2024.
\bibitem{b29} Z. Li, X. Yang, K. Choi, W. Zhu, R. Hsieh, H. Kim, J. H. Lim, S. Ji, B. Lee, X. Yan, et al., ``Mmsci: A multimodal multi-discipline dataset for PhD-level scientific comprehension,'' in \textit{AI for Accelerated Materials Design-Vienna 2024}, 2024.
\bibitem{b30} M. Jiang, J. Gao, J. Zhan, and D. Wang, ``Mac: A live benchmark for multimodal large language models in scientific understanding,'' \textit{arXiv preprint arXiv:2508.15802}, 2025.
\bibitem{b31} J. Ruan, D. Jiang, X. Gao, T. Liu, Y. Fu, and Y. Kang, ``Mme-sci: A comprehensive and challenging science benchmark for multimodal large language models,'' in \textit{Proceedings of the AAAI Conference on Artificial Intelligence}, vol. 40, no. 11, pp. 8760--8768, 2026.
\bibitem{b32} J. Ai, P. Zhou, Z. Xu, M. Li, F. Zhang, Z. Li, J. Sun, Y. Feng, B. Huang, Z. Wang, et al., ``Projudge: A multi-modal multi-discipline benchmark and instruction-tuning dataset for MLLM-based process judges,'' in \textit{Proceedings of the IEEE/CVF International Conference on Computer Vision}, pp. 4681--4690, 2025.
\bibitem{b33} A. Mukherjee and S. Ghosh, ``mmJEE-Eval: A bilingual multimodal benchmark for evaluating scientific reasoning in vision-language models,'' in \textit{Proceedings of the 14th International Joint Conference on Natural Language Processing and the 4th Conference of the Asia-Pacific Chapter of the Association for Computational Linguistics}, pp. 2268--2290, 2025.
\bibitem{b34} P. Zhou, X. Peng, F. Zhang, Z. Xu, J. Ai, Y. Qiu, W. Zhao, J. Song, C. Li, W. Tang, et al., ``Mdk12-bench: A multi-discipline benchmark for evaluating reasoning in multimodal large language models,'' in \textit{Proceedings of the AAAI Conference on Artificial Intelligence}, vol. 40, no. 34, pp. 28982--28990, 2026.
\bibitem{b35} X. Yue, Y. Ni, K. Zhang, T. Zheng, R. Liu, G. Zhang, S. Stevens, D. Jiang, W. Ren, Y. Sun, et al., ``Mmmu: A massive multi-discipline multimodal understanding and reasoning benchmark for expert AGI,'' in \textit{Proceedings of the IEEE/CVF Conference on Computer Vision and Pattern Recognition}, pp. 9556--9567, 2024.
\bibitem{b36} Springer, ``Volumes and issues,'' \textit{Journal of Phase Equilibria}, Springer Nature Link. [Online]. Available: https://link.springer.com/journal/12385/volumes-and-issues. [Accessed: Jun. 6, 2026].
\bibitem{b37} B. Wang, C. Xu, X. Zhao, L. Ouyang, F. Wu, Z. Zhao, R. Xu, K. Liu, Y. Qu, F. Shang, et al., ``Mineru: An open-source solution for precise document content extraction,'' \textit{arXiv preprint arXiv:2409.18839}, 2024. [Online]. Available: \url{https://arxiv.org/abs/2409.18839}
\bibitem{b38} Z.ai, ``GLM‑5.1,'' Z.ai Blog, Apr. 2026. [Online]. Available: https://z.ai/blog/glm-5.1
\bibitem{b39} K. L. Gwet, ``Computing inter-rater reliability and its variance in the presence of high agreement,'' \textit{British Journal of Mathematical and Statistical Psychology}, vol. 61, no. 1, pp. 29--48, 2008.
\bibitem{b40} J. Cohen, ``A coefficient of agreement for nominal scales,'' \textit{Educational and Psychological Measurement}, vol. 20, no. 1, pp. 37--46, 1960.
\bibitem{b41} C.-Y. Lin, ``Rouge: A package for automatic evaluation of summaries,'' in \textit{Text Summarization Branches Out}, pp. 74--81, 2004.
\bibitem{b42} T. Zhang, V. Kishore, F. Wu, K. Q. Weinberger, and Y. Artzi, ``Bertscore: Evaluating text generation with BERT,'' \textit{arXiv preprint arXiv:1904.09675}, 2019.
\bibitem{b43} Z.AI, ``GLM‑5V‑Turbo,'' \textit{Z.AI Developer Documentation}, [Online]. Available: https://docs.z.ai/guides/vlm/glm-5v-turbo. [Accessed: Jun. 6, 2026].

\end{thebibliography}
\end{document}